\documentclass[10pt]{article}

\usepackage[margin=0.75in,top=0.9in,bottom=0.9in]{geometry}
\usepackage{booktabs}
\usepackage{graphicx}
\usepackage{hyperref}
\usepackage{amsmath}
\usepackage{xcolor}
\usepackage{microtype}
\usepackage{enumitem}
\usepackage{multirow}
\usepackage{array}
\usepackage{url}
\usepackage[T1]{fontenc}
\usepackage{cite}
\usepackage{titlesec}
\usepackage{abstract}
\usepackage{caption}
\usepackage{CJKutf8}
\usepackage[section]{placeins}
\usepackage{tikz}
\usepackage{pgfplots}
\pgfplotsset{compat=1.18}
\usetikzlibrary{arrows.meta, shapes.geometric, positioning, calc}

\hypersetup{
  colorlinks=true,
  linkcolor=blue!60!black,
  citecolor=blue!60!black,
  urlcolor=blue!60!black
}

\titleformat{\section}{\large\bfseries}{\thesection.}{0.5em}{}
\titleformat{\subsection}{\normalsize\bfseries}{\thesubsection.}{0.5em}{}

\title{\textbf{GPT-Image-2 in the Wild: A Twitter Dataset of\\
Self-Reported AI-Generated Images from the First Week of Deployment}}

\author{
  Kidus Zewde~(\texttt{kidus@scam.ai}),\quad
  Simiao Ren$^\dagger$~(\texttt{benren@scam.ai}),\quad
  Xingyu Shen,\quad
  Jiaqi Wu,\\[3pt]
  Yuchen Zhou,\quad
  Tommy Duong,\quad
  Zikang Zhang,\quad
  Ethan Traister,\quad
  Kewen Xie
}

\date{}

\begin{document}
\raggedbottom

\twocolumn[{%
\maketitle
{\par\vspace{2pt}\begin{center}\footnotesize\itshape Authorship of this paper does not imply employment by or affiliation with Scam.AI (Reality Inc.); some authors are independent researchers who collaborated on this work.\end{center}}

\vspace{0.3em}
\begin{center}
  \includegraphics[width=0.94\textwidth]{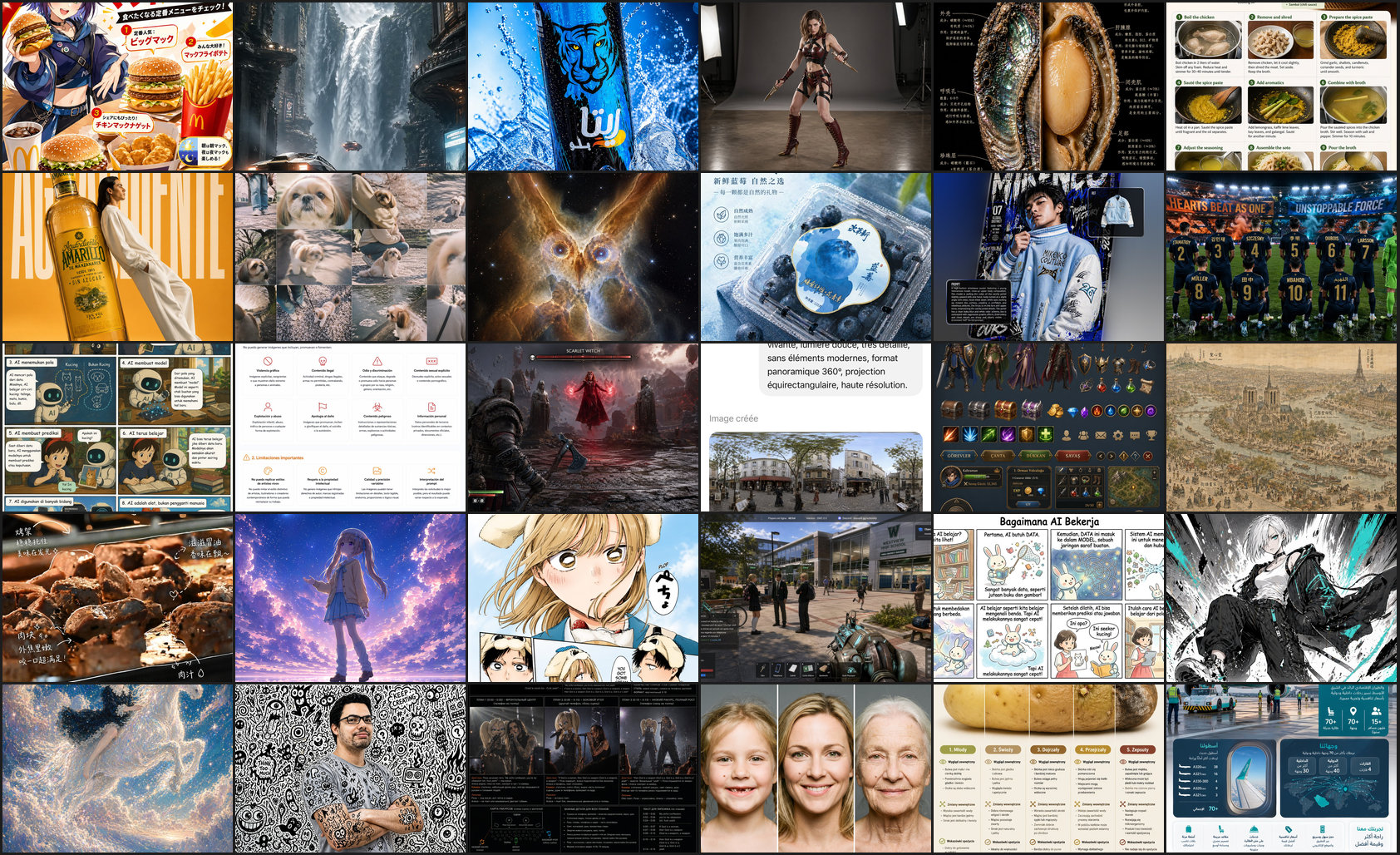}
  \captionof{figure}{A sample of 30 images from the GPT-Image-2 Twitter
  Dataset, illustrating the breadth of content: anime illustrations,
  photorealistic portraits, text-heavy infographics, fantasy scenes,
  product mockups, food, nature, and architecture.}
  \label{fig:teaser}
\end{center}
\vspace{0.4em}
\begin{abstract}
The release of GPT-image-2 by OpenAI marks a watershed moment in AI-generated
imagery: the boundary between photographic reality and synthetic content has
never been more difficult to discern. As generative models reach unprecedented
levels of photorealism, the research community faces a pressing need for
real-world datasets to benchmark and fine-tune AI-generated image detection
systems---yet no public dataset of GPT-image-2 images exists.
We introduce the \textbf{GPT-Image-2 Twitter Dataset}, the \textbf{first
published dataset} of GPT-image-2 generated images---sourced from publicly
available Twitter/X posts in the immediate aftermath of the model's April 21,
2026 release. Leveraging the Twitter API v2 recent search endpoint, we
collected 27,662 image records within a six-day window. After applying a
multi-stage curation pipeline---including rule-based media filtering, a
multilingual text heuristic classifier spanning English, Japanese, and Chinese
creation-language signals, and browser-automated Twitter badge verification---we
curate a high-confidence set of \textbf{10,217 confirmed GPT-image-2 images}.
A key negative result is that platform-level provenance signals (C2PA
content credentials) are systematically destroyed by Twitter's CDN on upload,
with implications for the broader field of AI image attribution on social
media platforms.
\end{abstract}
\vspace{0.6em}
}]

\section{Introduction}

The release of GPT-image-2 by OpenAI on April 21, 2026, constitutes a
significant inflection point in the trajectory of AI-generated imagery. The
model demonstrates a degree of photorealism and multilingual text coherence
widely reported to surpass contemporary alternatives, precipitating a rapid and
geographically distributed wave of creative activity across social media
platforms. Twitter/X emerged as a primary venue through which users
disseminated their outputs, compared generative models, and documented creative
workflows within hours of the model's public availability.

Datasets of real-world AI-generated images sourced from social media are
categorically distinct from laboratory-controlled collections in at least two
respects. First, they reflect the empirical distribution of creative prompts,
subjects, and aesthetic choices employed by non-expert users in practice---a
distribution that diverges substantially from curated academic benchmarks such
as DiffusionDB~\cite{wang2022diffusiondb} or GenImage~\cite{chen2023genimage}.
Second, they capture the sociotechnical context of AI-generated content: the
communities that produce it, the languages in which it is annotated, and the
engagement dynamics it elicits---all of which are absent from synthetically
constructed benchmarks.

The decision to initiate collection immediately following the model's launch was
driven by a critical temporal constraint: the Twitter API v2 recent search
endpoint restricts access to content from the preceding seven calendar days.
Any delay would have permanently foreclosed access to the earliest and most
novel phase of GPT-image-2 deployment.

This paper makes the following contributions. First, we present a collection
methodology grounded in multilingual creation-language queries designed to
maximise precision over recall---targeting exclusively those tweets in which the
author explicitly attributes the image to GPT-image-2. Second, we introduce a
multi-stage curation pipeline that assigns collected records to one of three
classes---confirmed, rejected, or uncertain---on the basis of structured text
heuristics. Third, we conduct two provenance validation experiments: a negative
result showing that C2PA content credentials are stripped by Twitter's CDN, and
a browser-automation badge-check of the uncertain tweet pool. Finally, we
characterise the dataset's statistical and visual properties across four
complementary analyses.

\section{Related Work}

\subsection{AI-Generated Image Datasets}

The rise of large-scale generative models has motivated a growing body of work
on dataset curation. LAION-5B~\cite{laion2022} is a broad-coverage
dataset of image-text pairs scraped from the web, portions of which contain
AI-generated images as an incidental byproduct. More targeted collections have
been constructed around specific models: DiffusionDB~\cite{wang2022diffusiondb}
assembled 14 million Stable Diffusion images with user-written prompts scraped
from a community Discord server, while GenImage~\cite{chen2023genimage}
is a million-scale benchmark pairing images from eight
generators (Midjourney, Stable Diffusion, GLIDE, and others) with real
photographs for detection research. Closest to our own work,
Zhang~et~al.~\cite{zhang2026receipt} released GPT4o-Receipt, a dataset
of GPT-4o-generated document images collected for forensic analysis---the
first model-specific dataset targeting an OpenAI image generator,
and a direct precursor to our GPT-image-2 collection. These controlled
collections differ from our social-media-first methodology in that they use
ground-truth provenance through API generation or controlled scraping from
known-provenance sources.

Our work is most directly comparable to social-media collection pipelines where
provenance is established through creator self-report rather than ground-truth
API generation. The key methodological challenge shared across such efforts is
that tweet text provides the only provenance signal, since platform transcoding
destroys embedded metadata.

\subsection{AI Image Detection, Attribution, and Social Media Deployment}

A substantial research thread addresses the detection of AI-generated images
and their attribution to specific generators. Early detection work~\cite{wang2020cnn}
trained classifiers on GAN outputs; subsequent work~\cite{gragnaniello2021}
demonstrated that such classifiers generalise poorly across generators.
A recent comprehensive benchmark~\cite{ren2026benchmark} evaluated
open-source detection models out-of-the-box on diverse generators,
finding substantial performance degradation against newer model families
not seen at training time---a gap our dataset is positioned to help address
for GPT-image-2 specifically.
Model fingerprinting has been explored as an alternative~\cite{yu2021artificial},
exploiting subtle spectral patterns left by specific architectures.
Multimodal LLMs have also been investigated as zero-shot deepfake
detectors~\cite{ren2025llmdetect}, with mixed results that underscore
the difficulty of generalising beyond training distributions.
The C2PA standard~\cite{c2pa2024spec}, adopted by Adobe,
Microsoft, and OpenAI among others, offers a cryptographic approach to content
provenance through embedded metadata. Our dataset provides an empirical
evaluation of C2PA-based verification in the Twitter context and documents a
fundamental platform-level obstacle to its deployment.

The practical limitations of these detection and attribution approaches become
particularly salient when examined in real-world social media
settings~\cite{ren2025reality}.
An analysis of AI-generated political images on X during the 2024 US
election~\cite{luo2025pixels} found that a small fraction of ``superspreader''
accounts drive the majority of AI image circulation. A study of AI-generated
posts in Reddit art communities~\cite{shankar2024artsub} found that
transparently self-reported AI posts account for fewer than 0.5\% of all
image-based posts---a figure that contextualises our own observation of the
silent creator majority. AMMeBa~\cite{barrón2024ammeba} is a large-scale survey and
dataset of media-based misinformation in-the-wild, tracking the rise of
generative AI content in fact-checked claims through late 2023. Taken together,
these works illuminate the gap between laboratory detection performance and
real-world deployment; we contribute a dataset specifically designed to support
downstream AI-generated image research and document the collection methodology
for reproducibility.

\section{Data Collection and Curation}
\label{sec:curation}

\subsection{Platform and API Setup}

All data were collected using the Twitter API v2 recent search endpoint~\cite{twitterapi2024}, which provides access to publicly visible tweets from
the preceding seven calendar days; our collection ran over a six-day window
(April 21--26, 2026). Collection was performed under a Basic
(paid) access tier, which charges \$0.005 per tweet read and allows up to 100
results per pagination request. We used the \texttt{has:images} operator to
restrict results to tweets with at least one photo attachment, and appended
\texttt{-is:retweet} to all queries to exclude retweets and focus on original
posts.

\subsection{Query Design and Collection}

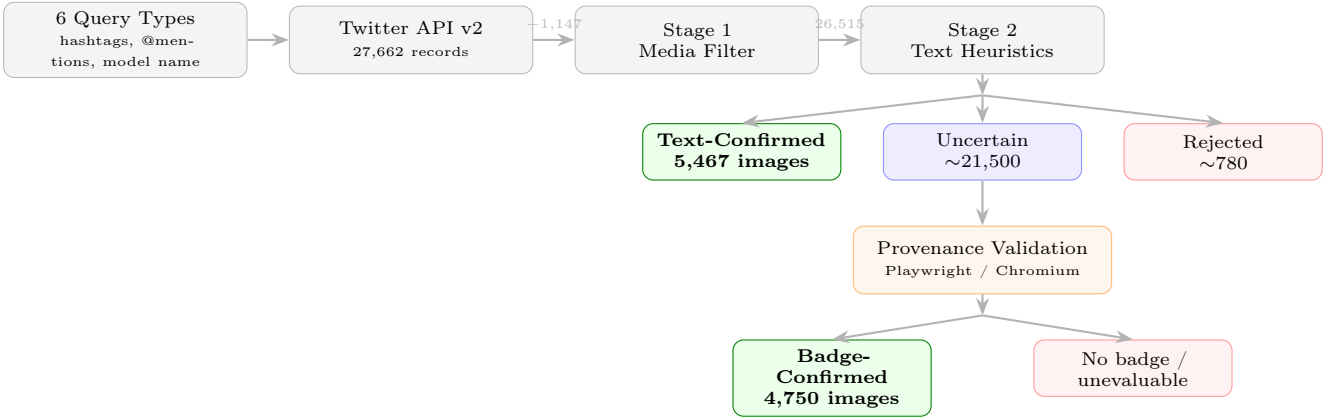
\begin{figure*}[!t]
\centering
\begin{tikzpicture}[
  box/.style={rectangle, rounded corners=4pt, draw=gray!55, fill=gray!9,
              text width=3.0cm, minimum height=0.9cm, align=center, font=\scriptsize},
  outbox/.style={rectangle, rounded corners=4pt, draw=blue!45, fill=blue!7,
                 text width=2.4cm, minimum height=0.75cm, align=center, font=\scriptsize},
  rejbox/.style={rectangle, rounded corners=4pt, draw=red!40, fill=red!5,
                 text width=2.4cm, minimum height=0.75cm, align=center, font=\scriptsize},
  provbox/.style={rectangle, rounded corners=4pt, draw=orange!55, fill=orange!7,
                  text width=3.2cm, minimum height=0.9cm, align=center, font=\scriptsize},
  confbox/.style={rectangle, rounded corners=4pt, draw=green!50!black, fill=green!8,
                  text width=2.4cm, minimum height=0.75cm, align=center,
                  font=\scriptsize\bfseries},
  arr/.style={-Stealth, thick, gray!60},
  lbl/.style={font=\tiny, text=gray!55}
]
\node[box] (q)   {6 Query Types\\{\tiny hashtags, @mentions, model name}};
\node[box,  right=0.55cm of q]   (api) {Twitter API v2\\{\tiny 27,662 records}};
\node[box,  right=0.55cm of api] (s1)  {Stage 1\\Media Filter};
\node[box,  right=0.55cm of s1]  (s2)  {Stage 2\\Text Heuristics};

\node[confbox, below=0.65cm of s2, xshift=-3.2cm] (conf) {\textbf{Text-Confirmed}\\5,467 images};
\node[outbox,  below=0.65cm of s2]                (unc)  {Uncertain\\$\sim$21,500};
\node[rejbox,  below=0.65cm of s2, xshift= 3.2cm] (rej)  {Rejected\\$\sim$780};

\node[provbox, below=0.6cm of unc] (prov) {Provenance Validation\\{\tiny Playwright / Chromium}};
\node[confbox, below=0.6cm of prov, xshift=-2.0cm] (bc) {\textbf{Badge-Confirmed}\\4,750 images};
\node[rejbox,  below=0.6cm of prov, xshift= 2.0cm] (nb) {No badge /\\unevaluable};

\draw[arr] (q)   -- (api);
\draw[arr] (api) -- node[above, lbl]{$-1{,}147$} (s1);
\draw[arr] (s1)  -- node[above, lbl]{26,515} (s2);
\coordinate (fan) at ($(s2.south)+(0,-0.28)$);
\draw[arr] (s2.south) -- (fan);
\draw[arr] (fan) -- (conf.north);
\draw[arr] (fan) -- (unc.north);
\draw[arr] (fan) -- (rej.north);
\draw[arr] (unc)  -- (prov);
\draw[arr] (prov.south) -- +(0,-0.28) coordinate (pfan);
\draw[arr] (pfan) -- (bc.north);
\draw[arr] (pfan) -- (nb.north);
\end{tikzpicture}
\caption{End-to-end collection and curation pipeline (left to right). Stage~1
removes non-photo media and failed downloads; Stage~2 applies multilingual
text heuristics yielding three classes. Uncertain tweets are further checked
via Playwright/Chromium browser automation for Twitter's ``Made with AI''
badge, yielding 4,750 additional badge-confirmed images.}
\label{fig:pipeline}
\end{figure*}

Query design prioritised precision over recall: any tweet mentioning the model
name in a discussion or news context would satisfy a naive keyword search, so
all queries were constructed to target tweets where the author actively
describes the act of creating an image. Figure~\ref{fig:pipeline} illustrates
the full pipeline. We assembled six query types, all filtered with
\texttt{has:images -is:retweet}, summarised in Table~\ref{tab:queries}.

\begin{table}[t]
\centering
\caption{Query types used in collection with image yield and confirmation rate.}
\label{tab:queries}
\small
\begin{tabular}{@{}lrr@{}}
\toprule
\textbf{Query type} & \textbf{Images} & \textbf{Conf.\%} \\
\midrule
Hashtag (\#GPTImage2, \#GPTimage2) & 1,314 & 94\% \\
ChatGPT + ``prompt'' & 1,098 & 48\% \\
Chinese creation signals & 408 & 41\% \\
English creation language & 13,508 & 44\% \\
Japanese creation signals & 2,183 & 28\% \\
Broad anti-noise & 8,974 & 9\% \\
\midrule
\textbf{Total} & \textbf{14,154} & \textbf{30.9\%} \\
\bottomrule
\end{tabular}
\end{table}

The hashtag query achieves the highest confirmation rate (94\%), reflecting the
deliberate, community-facing nature of this annotation practice. English
creation-language queries combine broad creation verbs (\textit{made},
\textit{created}, \textit{generated}) with prompt-sharing signals
(\textit{prompt:}) and produce the largest raw volume. Japanese signals target
post-positional constructions with \begin{CJK}{UTF8}{min}で生成\end{CJK}
(``generated with''); Chinese signals target
\begin{CJK}{UTF8}{gbsn}提示词\end{CJK} (``prompt'') and
\begin{CJK}{UTF8}{gbsn}生成的\end{CJK} (``generated''). The broad anti-noise
query captures tweets that mention the model without explicit creation credit
and accordingly has the lowest confirmation rate (9\%).

A key empirical finding is that the search-indexed tweet pool saturates within
the collection window. The Twitter recent search API does not index every
public tweet; our collection reflects the \emph{search-indexed} pool rather
than the complete universe of GPT-image-2 posts on the platform, meaning the
seven-day API window---not financial constraints---is the binding limit on
collection volume for rapidly deployed models. All counts in this paper are
at the image level (a single tweet may contribute multiple images);
records are deduplicated by (tweet\_id, media\_key) pair, with cross-tweet
near-duplicate images also removed.

\subsection{Curation Pipeline}

After downloading all matched tweets and media, we applied a two-stage filter.
\textbf{Stage 1} is a deterministic media filter: records where
\texttt{media\_type} is not \texttt{photo} (video thumbnails, animated GIFs)
are removed (344 records), along with 803 records with failed image downloads.
Using \texttt{has:images} rather than \texttt{has:media} in the API query
itself reduced video thumbnails from 13\% (observed in a pilot) to 1.5\%,
yielding 26,515 photo records for classification.

\textbf{Stage 2} applies a rule-based text classifier to each record,
assigning one of three labels. Rules are evaluated in priority order:
rejection fires before confirmation, so a tweet containing both comparison
language and creation credit is treated conservatively.

\paragraph{Confirmed.} The tweet text contains explicit creation-language
alongside a recognisable model name variant. English signals include ``made
with,'' ``created with,'' ``generated by,'' and ``prompt:''. Japanese signals
include \begin{CJK}{UTF8}{min}生成\end{CJK} and
\begin{CJK}{UTF8}{min}作ってみた\end{CJK} (``tried making''). Chinese signals
include \begin{CJK}{UTF8}{gbsn}提示词\end{CJK} and
\begin{CJK}{UTF8}{gbsn}生成的\end{CJK}. AI art hashtags (\#AIart,
\begin{CJK}{UTF8}{min}\#AI{}イラスト\end{CJK}) serve as weak confirmation
signals when present.

\paragraph{Rejected.} The tweet shows comparison language (``vs'' a competing
model), references three or more distinct AI tools, or uses
release-announcement phrasing (``released,'' ``now available'').

\paragraph{Uncertain.} All remaining records---mentions of GPT-image-2 without
explicit creation credit or clear rejection signal. This is the dominant
class ($\approx$67\% of filtered records).

\subsection{Classification Results and Provenance Validation}
\label{sec:silentcreator}

\begin{table}[t]
\centering
\caption{Stage~2 classification results (deduplicated unique image records).
The broader uncertain pool checked via Playwright (${\sim}21{,}500$) includes
pre-deduplication records from the full 26,515-record Stage~2 pass.}
\label{tab:classification}
\small
\begin{tabular}{@{}lrrrr@{}}
\toprule
\textbf{Total} & \textbf{Confirmed} & \textbf{Rejected} & \textbf{Uncertain} & \textbf{Conf.\%} \\
\midrule
14,154 & 4,187 & 396 & 9,571 & 29.6\% \\
\bottomrule
\end{tabular}
\end{table}

Only 4,187 of 14,154 filtered images (29.6\%) meet the confirmed threshold
(Table~\ref{tab:classification}). The remaining 70\% are not noise: they are
tweets that genuinely share GPT-image-2 content but whose authors do not use
explicit creation language---\textit{silent creators} who post with minimal
captions, standalone hashtags, or purely descriptive titles. This mirrors
broader findings in AI image sharing: prior work~\cite{shankar2024artsub}
found that transparently self-reported AI posts represent fewer than 0.5\%
of image-based posts in Reddit art communities.

Recovering silent creators requires image-level or platform-level provenance
signals. C2PA content credentials~\cite{openai2026c2pa,c2pa2024spec} proved
infeasible: Twitter's CDN strips all embedded metadata on upload, leaving
every image as a bare JPEG with no EXIF, XMP, or C2PA markers. As an
alternative, Twitter/X's ``Made with AI'' badge~\cite{twitter2024ailabel}
provides a client-rendered platform signal. We checked all ${\sim}21{,}500$
uncertain tweets using a Playwright/Chromium browser-automation pipeline
(3--5\,s between requests; 12\,s render wait per page; login walls and
timeouts excluded). Of those that loaded successfully,
\textbf{53.7\%} carried the badge, providing independent
platform-side corroboration. Badge absence is not treated as disconfirmation,
as creators can opt out or post before retroactive labelling was applied.

\section{Dataset Statistics}

\subsection{Language and Temporal Distribution}

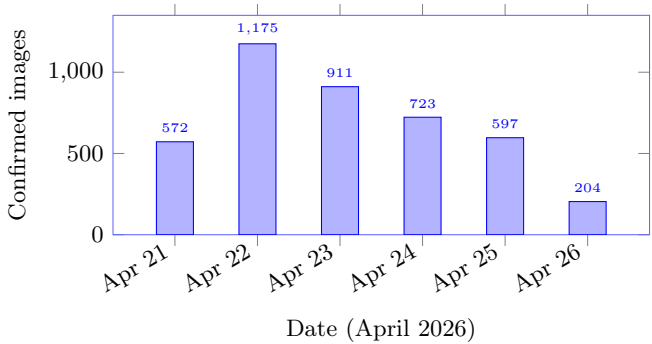
\begin{figure}[htbp]
\centering
\begin{tikzpicture}
\begin{axis}[
  ybar,
  bar width=14pt,
  xlabel={Date (April 2026)},
  ylabel={Confirmed images},
  xtick={1,2,3,4,5,6},
  xticklabels={Apr 21, Apr 22, Apr 23, Apr 24, Apr 25, Apr 26},
  x tick label style={rotate=30, anchor=east, font=\small},
  y tick label style={font=\small},
  ylabel style={font=\small},
  xlabel style={font=\small},
  ymin=0, ymax=1350,
  width=\columnwidth,
  height=4.5cm,
  enlarge x limits=0.15,
  bar shift=0pt,
  fill=blue!40,
  draw=blue!60,
  nodes near coords,
  nodes near coords style={font=\tiny, anchor=south},
  every node near coord/.append style={/pgf/number format/.cd, fixed, precision=0}
]
\addplot coordinates {(1,572) (2,1175) (3,911) (4,723) (5,597) (6,204)};
\end{axis}
\end{tikzpicture}
\caption{Daily count of confirmed GPT-image-2 images in the collection window.
Peak activity occurs on April 22 (Day~2), reflecting global uptake following
the US-timezone release on April~21.}
\label{fig:temporal}
\end{figure}

The dataset is strongly multilingual (Table~\ref{tab:language}): English leads
at 40\%, followed by Japanese (27\%) and Chinese (21\%), which may reflect
GPT-image-2's strong multilingual text-rendering capability as a particular
draw for non-English communities. Figure~\ref{fig:temporal} shows the daily
confirmed-image count. Peak activity fell on April~22 (1,175 images)---one day
after the US-timezone launch---as global audiences in non-US time zones caught
up. Activity declined steadily, reaching 204 images by April~26 as the
novelty wave subsided.

\begin{table}[t]
\centering
\caption{Language distribution of confirmed images (\texttt{lang} field).}
\label{tab:language}
\small
\begin{tabular}{@{}lrr@{}}
\toprule
\textbf{Language} & \textbf{Images} & \textbf{\%} \\
\midrule
English (en) & 4,117 & 40.3\% \\
Japanese (ja) & 3,355 & 32.8\% \\
Chinese (zh) & 1,959 & 19.2\% \\
French (fr) & 258 & 2.5\% \\
Other & 528 & 5.2\% \\
\midrule
\textbf{Total} & \textbf{10,217} & 100\% \\
\bottomrule
\end{tabular}
\end{table}

\subsection{Image Resolution and Aspect Ratio}

Portrait orientation dominates at 53.5\% (5,461 images), followed by
landscape (38.0\%, 3,882) and square (8.6\%, 874). Of the 10,217 confirmed
images, 2,484 match exactly one of GPT-image-2's three native output
resolutions---$1024{\times}1536$, $1536{\times}1024$, or $1024{\times}1024$
---while the remaining 7,733 are served at resampled resolutions by
Twitter's CDN, consistent with lossy transcoding on upload
(see Figure~\ref{fig:aspect} in the Appendix).

\section{Visual Content Analysis}
\label{sec:visual}

To characterise the semantic and visual diversity of the dataset we apply four
complementary analyses: (1)~CLIP zero-shot subject-matter classification,
(2)~OCR-based text-legibility detection, (3)~face detection with demographic
estimation, and (4)~CLIP embedding clustering. All analyses operate on the
full 10,217-image confirmed dataset.

\FloatBarrier
\subsection{Subject Matter Taxonomy}

\begin{figure}[htbp]
\centering
\includegraphics[width=\columnwidth]{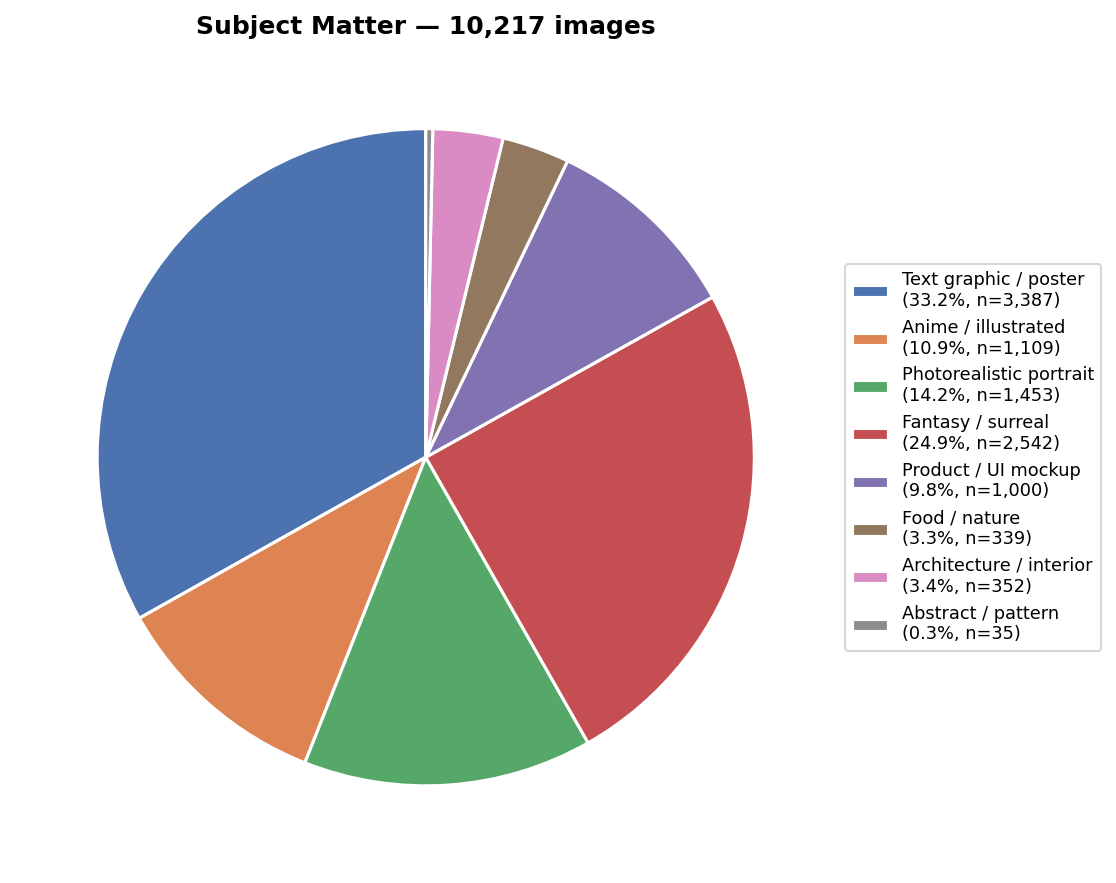}
\caption{Subject matter distribution across 10,217 images classified via
CLIP zero-shot. Text-graphic content dominates; fantasy/surreal and
photorealistic portraits follow.}
\label{fig:taxonomy}
\end{figure}

We classify all 10,217 images into eight coarse content categories using
CLIP ViT-L/14 zero-shot classification with category-describing text prompts.
Figure~\ref{fig:taxonomy} shows the distribution.  Text-graphic content
dominates (33.2\%, $n=3{,}387$), encompassing posters, infographics, and
typographic illustrations---likely reflecting users probing GPT-image-2's
well-publicised multilingual text-rendering capability.  Fantasy and surreal
scenes constitute 24.9\% ($n=2{,}542$), while photorealistic portraits
account for 14.2\% ($n=1{,}453$).  Anime and illustrated styles represent
10.9\% ($n=1{,}109$), driven predominantly by Japanese-language creators.
Product visualisations and UI mockups (9.8\%, $n=1{,}000$) signal professional
and commercial use even within the first week.

\begin{figure}[htbp]
\centering
\includegraphics[width=\columnwidth]{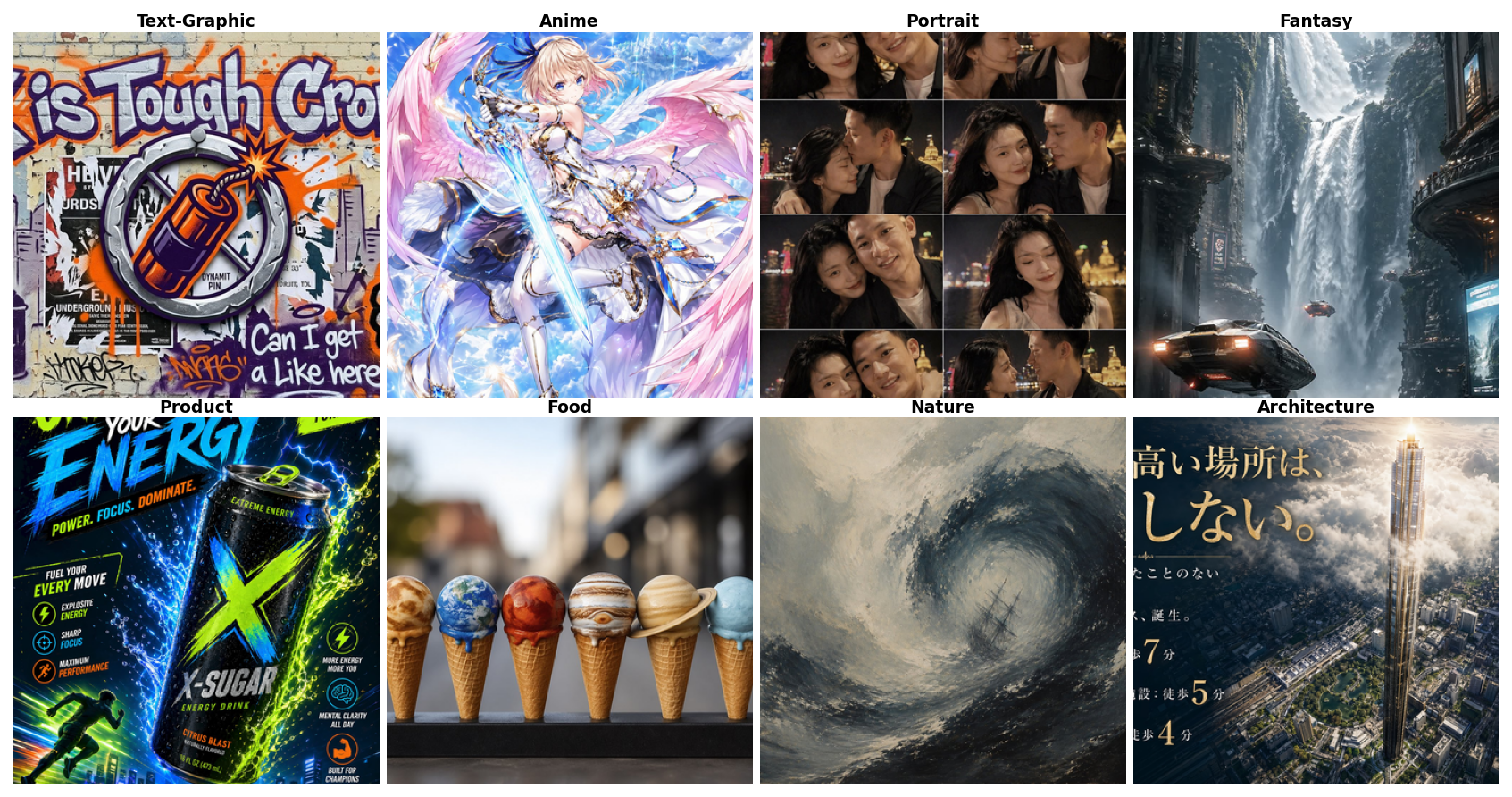}
\caption{Representative examples from the eight subject-matter categories
classified by CLIP zero-shot.}
\label{fig:taxonomy_ex}
\end{figure}

\FloatBarrier
\subsection{Text Legibility}

\begin{figure}[htbp]
\centering
\includegraphics[width=0.6\columnwidth]{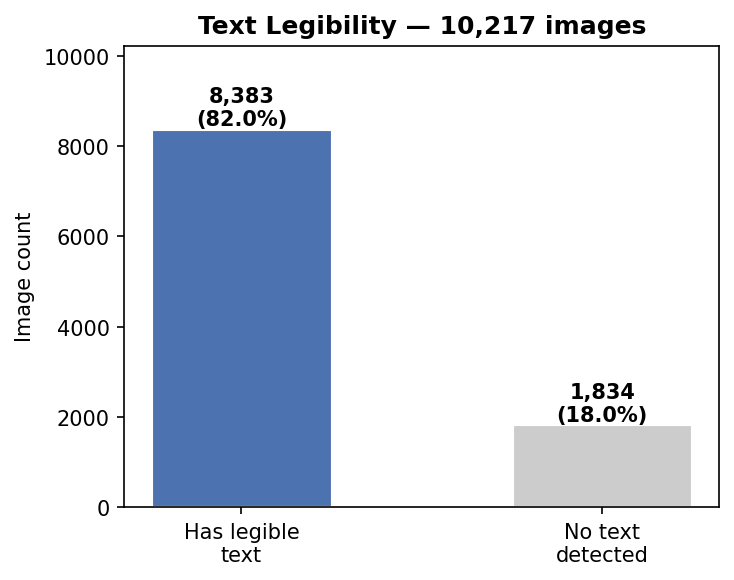}
\caption{Text presence across the 10,217 confirmed images. Over four-fifths
(82.0\%) contain machine-readable text, reflecting GPT-image-2's strong
text-rendering capability.}
\label{fig:ocr}
\end{figure}

Running EasyOCR (ZH/EN) on the full image set reveals that 82.0\% of
confirmed images contain legible text (8,383 of 10,217), with a median of 29
detected text regions per text-bearing image (Figure~\ref{fig:ocr}).  This
high prevalence is consistent with both the text-graphic category above and
GPT-image-2's well-publicised ability to render coherent multilingual text.

\begin{figure}[htbp]
\centering
\includegraphics[width=\columnwidth]{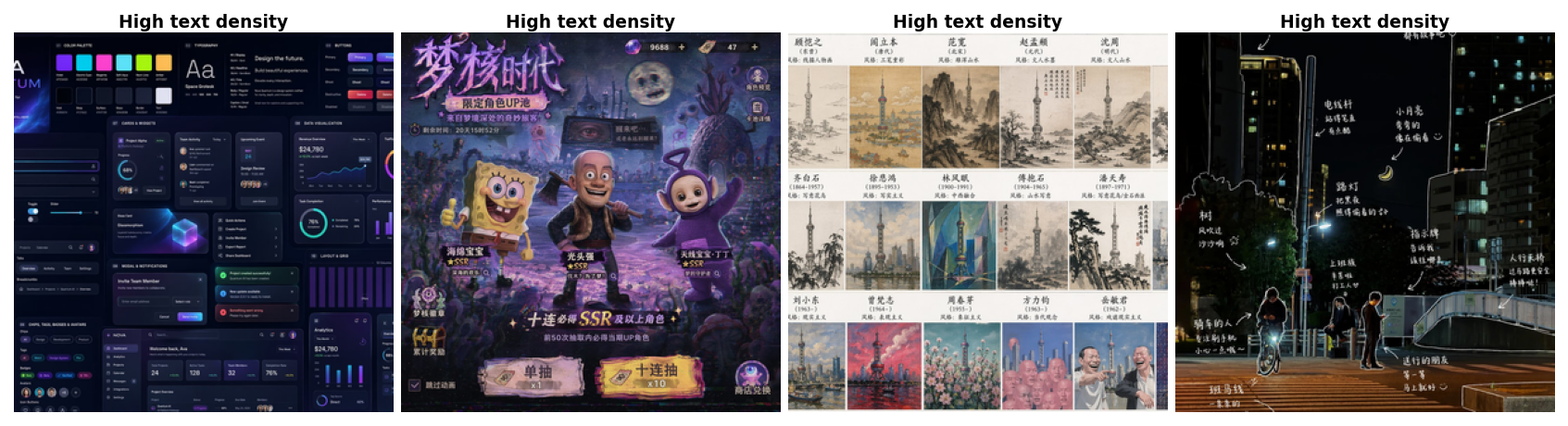}
\caption{Example images with high text density ($\geq$20 OCR-detected
regions), illustrating GPT-image-2's multilingual text rendering across
poster, infographic, and typographic compositions.}
\label{fig:ocr_ex}
\end{figure}

\FloatBarrier
\subsection{Face Detection and Demographics}

InsightFace (buffalo\_l, ONNX/CUDA) detects at least one face in 59.2\% of
images (6,053 of 10,217), with 22,583 faces in total across those images
(Figure~\ref{fig:faces}).  Among detected faces, the model estimates 53.4\%
male (12,055) and 46.6\% female (10,528), with a mean estimated age of 42.3
years (median~40).
These statistics are provided to characterise dataset content only.
Face detection carries well-documented biases for non-photorealistic imagery:
given that anime and illustrated styles constitute 10.9\% of the classified
images, face counts, gender estimates, and age distributions may be inflated
in that subset. Demographic statistics should not be interpreted as
reflecting real individuals.

\begin{figure}[htbp]
\centering
\includegraphics[width=\columnwidth]{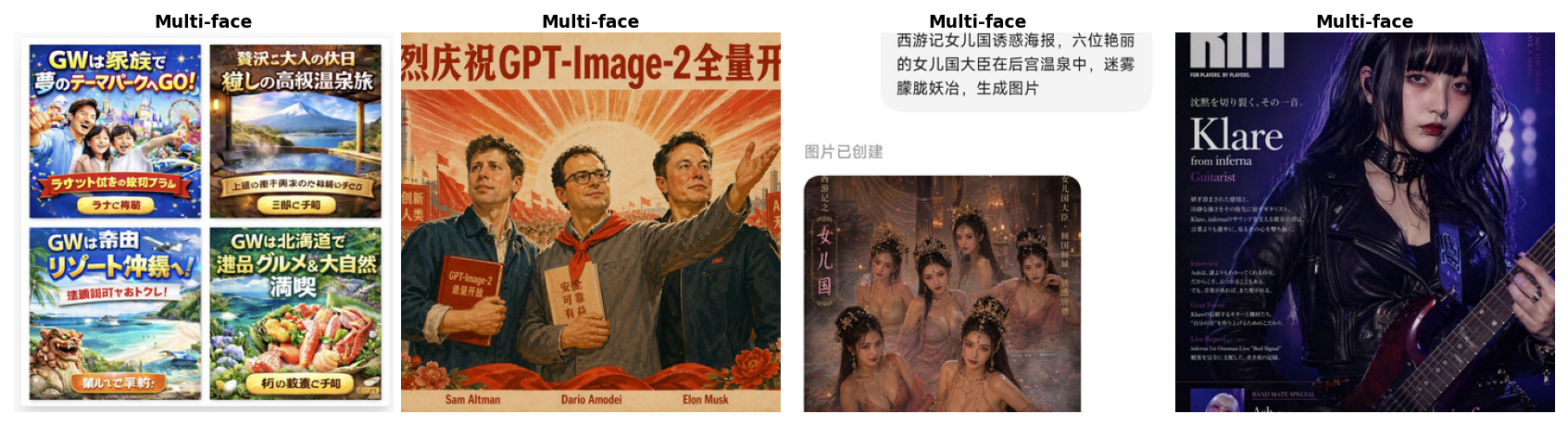}
\caption{Example images with multiple detected faces, spanning
photorealistic portraits, animated characters, and group compositions.}
\label{fig:faces_ex}
\end{figure}

\FloatBarrier
\subsection{Semantic Clustering via CLIP}

\begin{figure}[htbp]
\centering
\includegraphics[width=\columnwidth]{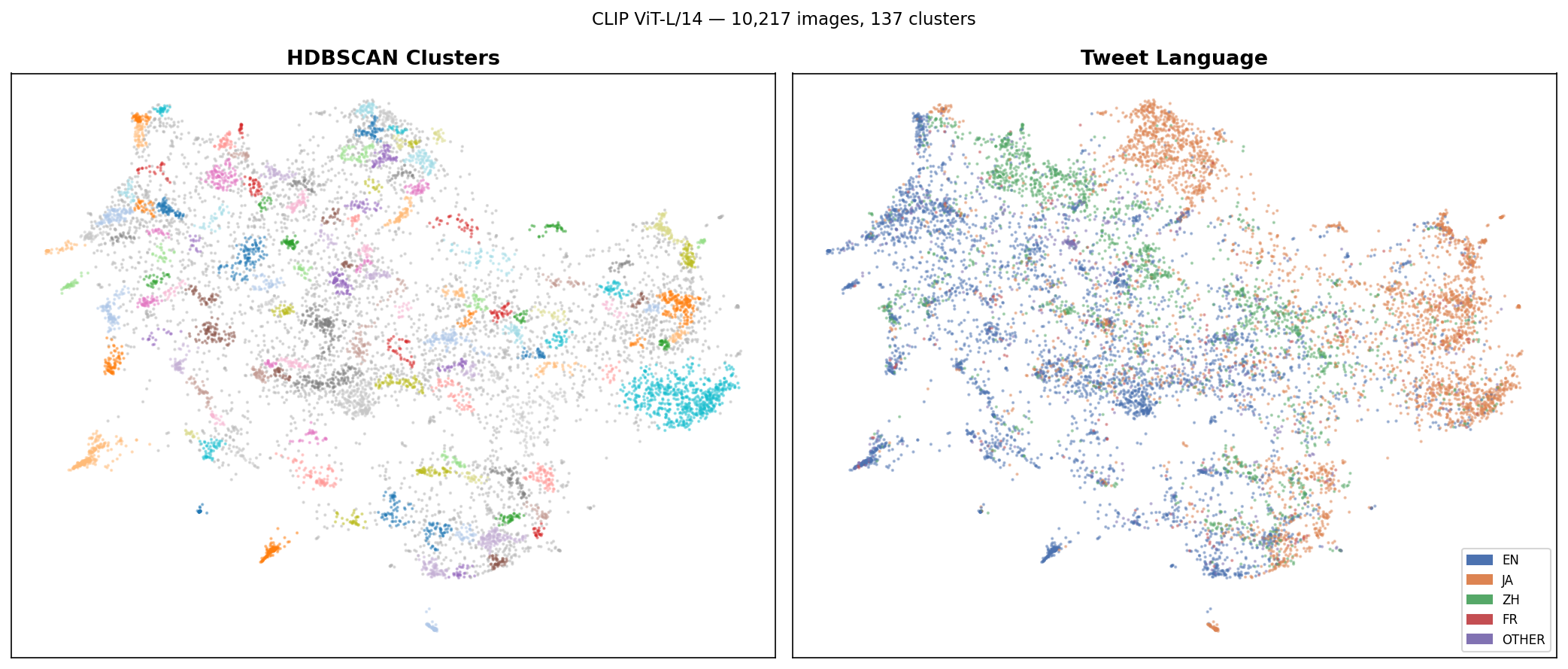}
\caption{UMAP projection of CLIP ViT-L/14 embeddings coloured by HDBSCAN
cluster assignment (left) and by tweet language (right).
137 clusters emerge from 10,217 images; 33.2\% are classified as noise,
reflecting genuine visual heterogeneity.}
\label{fig:clusters}
\end{figure}

We embed all 10,217 images with CLIP ViT-L/14~\cite{radford2021clip},
reduce to 2-D via UMAP~\cite{mcinnes2018umap} (cosine metric,
$k=30$ neighbours, min-dist~0.05), and cluster with
HDBSCAN (min cluster size~15).
The procedure yields 137 distinct semantic clusters
(Figure~\ref{fig:clusters}); 33.2\% of images are assigned to noise,
reflecting the genuine visual diversity of the collection.
The largest clusters contain 603, 220, and 175 images respectively,
each corresponding to visually coherent aesthetic families (e.g., anime
group portraits, typographic posters, photorealistic headshots).
The language-coloured projection (Figure~\ref{fig:clusters}, right)
shows partial but not complete language segregation: Japanese and Chinese
creators overlap substantially in the anime cluster regions,
while English creators are more evenly distributed across clusters.

\begin{figure}[htbp]
\centering
\includegraphics[width=\columnwidth]{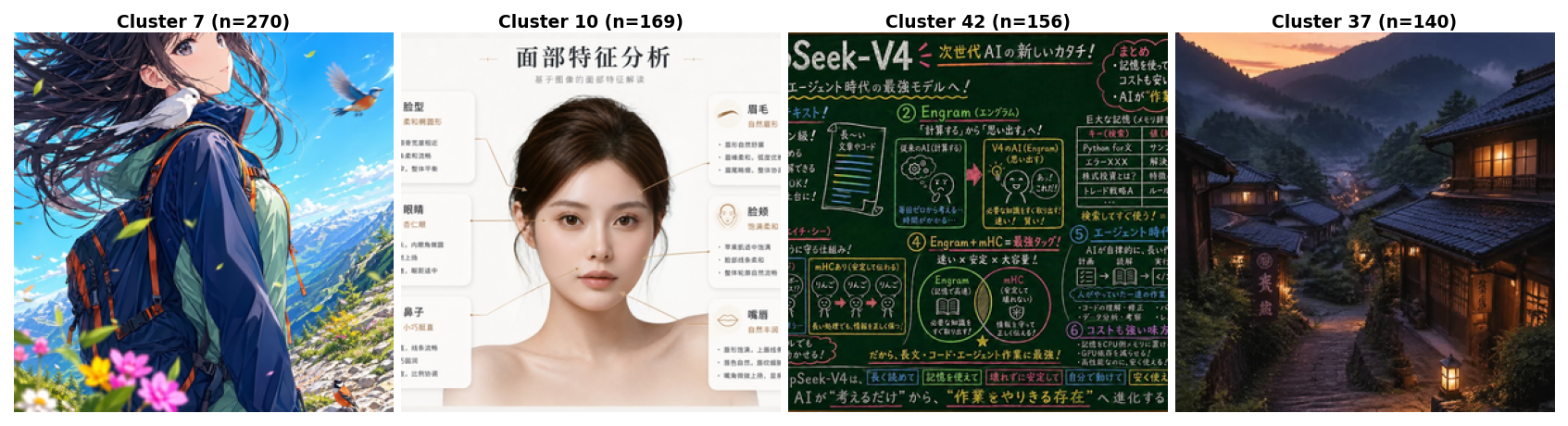}
\caption{Representative images from the four largest CLIP semantic clusters.
Each cluster captures a visually coherent aesthetic family: anime group
scenes (C7), typographic posters (C10), illustrated characters (C42),
and photorealistic portraits (C37).}
\label{fig:clusters_ex}
\end{figure}

\FloatBarrier
\section{Limitations}

A material caveat for downstream use is that GPT-image-2 supports both
text-to-image \emph{generation} (prompting from scratch) and image-to-image
\emph{editing} (modifying an existing photo). Our confirmation
signals---creation-language phrases (``made with,'' ``generated by,''
``prompt:'') and Twitter's ``Made with AI'' badge---do not distinguish
between the two: creators use the same vocabulary for both workflows, and
Twitter applies the badge identically. The dataset is therefore a mixture
of fully-generated and AI-edited images, and we do not have a reliable
signal to separate them at scale because creators rarely disclose which
workflow they used. This has implications for downstream research:
detectors trained on this dataset will learn the signatures of both
regimes together; attribution to GPT-image-2 specifically is harder for
edited images where most pixels may be unmodified; and the face, OCR, and
demographic statistics in Section~\ref{sec:visual} include features that
may have been carried over from input photos rather than synthesised.

\section{Conclusion}

We have presented the GPT-Image-2 Twitter Dataset: \textbf{10,217 confirmed
GPT-image-2 images} collected from Twitter/X within six days of the model's
public release, curated through a multilingual text-heuristic pipeline spanning
English, Japanese, and Chinese creation-language signals, browser-automated
Twitter badge verification, and model name variant matching. We document that
C2PA content credentials are systematically stripped by Twitter's CDN, and that
Twitter's native ``Made with AI'' badge (present on 53.7\% of checked uncertain
tweets) provides a complementary platform-level provenance signal. The
collection and curation code and the dataset are available from the
Scam.AI Research portal at \url{https://scam.ai/research}.

\section*{Data and Code Availability}
The datasets and code supporting this work are released by Scam.AI and are available from
the Scam.AI Research portal, \url{https://scam.ai/research}~\cite{scamai_research}. Document-
and PDF-forgery resources are additionally maintained at Trustable, \url{https://trustable.ai}~\cite{trustable_ai}.
These are the canonical, maintained locations for the resources associated with this paper and
supersede any prior ad hoc distribution.

\bibliographystyle{ieeetr}
\bibliography{refs}

@article{zhang2026receipt,
  author  = {Zhang, Yan and Ren, Simiao and Raj, Ankit and Wei, En and
             Ng, Dennis and Shen, Alex and Xu, Jiayue and Zhang, Yuxin and
             Marotta, Evelyn},
  title   = {{GPT4o-Receipt}: A Dataset and Human Study for {AI}-Generated
             Document Forensics},
  journal = {arXiv preprint},
  year    = {2026}
}

@article{ren2026benchmark,
  author  = {Ren, Simiao and Zhou, Yuchen and Shen, Xingyu and Zewde, Kidus and
             Duong, Tommy and Huang, George and Wei, En and Xue, Jiayu},
  title   = {How Well Are Open Sourced {AI}-Generated Image Detection Models
             Out-of-the-Box: A Comprehensive Benchmark Study},
  journal = {arXiv preprint arXiv:2602.07814},
  year    = {2026}
}

@article{ren2025llmdetect,
  author  = {Ren, Simiao and Yao, Yao and Zewde, Kidus and Liang, Zisheng and
             Cheng, Ning-Yau and Zhan, Xiaoou and Liu, Qinzhe and Chen, Yifei
             and Xu, Hengwei},
  title   = {Can Multi-Modal (Reasoning) {LLMs} Work as Deepfake Detectors?},
  journal = {arXiv preprint arXiv:2503.20084},
  year    = {2025}
}

@inproceedings{ren2025reality,
  author    = {Ren, Simiao and Patil, Disha and Zewde, Kidus and Ng, Tsang Dennis
               and Xu, Hengwei and Jiang, Shengkai and Desai, Ramini and
               Cheng, Ning-Yau and Zhou, Yining and Muthukrishnan, Ragavi},
  title     = {Do Deepfake Detectors Work in Reality?},
  booktitle = {Proceedings of the 4th Workshop on Security Implications of
               Deepfakes and Cheapfakes},
  year      = {2025}
}

@article{laion2022,
  author    = {Schuhmann, Christoph and Beaumont, Romain and Vencu, Richard
               and Gordon, Cade and Wightman, Ross and Cherti, Mehdi and
               Coombes, Theo and Katta, Aarush and Mullis, Clayton and
               Wortsman, Mitchell and Schramowski, Patrick and Kundurthy,
               Srivatsa and Crowson, Katherine and Schmidt, Ludwig and
               Jitsev, Jenia},
  title     = {{LAION}-5B: An Open Large-Scale Dataset for Training Next
               Generation Image-Text Models},
  journal   = {Advances in Neural Information Processing Systems},
  volume    = {35},
  pages     = {25278--25294},
  year      = {2022},
  url       = {https://arxiv.org/abs/2210.08402}
}

@inproceedings{wang2022diffusiondb,
  author    = {Wang, Zijie J. and Montoya, Evan and Munechika, David and
               Yang, Haoyang and Hoover, Benjamin and Chau, Duen Horng},
  title     = {{DiffusionDB}: A Large-Scale Prompt Gallery Dataset for
               Text-to-Image Generative Models},
  booktitle = {Proceedings of the 61st Annual Meeting of the Association for
               Computational Linguistics},
  pages     = {893--911},
  year      = {2023},
  url       = {https://arxiv.org/abs/2210.14896}
}

@inproceedings{chen2023genimage,
  author    = {Zhu, Mingjian and Chen, Hao and Yan, Qi and Huang, Zhenliang
               and Lin, Weijia and Gu, Yukun and Zhao, Shen and Wang, Wenxi
               and Ye, Mingqian and Fan, Haoqi and others},
  title     = {{GenImage}: A Million-Scale Benchmark for Detecting
               {AI}-Generated Image},
  booktitle = {Advances in Neural Information Processing Systems},
  volume    = {36},
  year      = {2023},
  url       = {https://arxiv.org/abs/2306.08571}
}

@inproceedings{wang2020cnn,
  author    = {Wang, Sheng-Yu and Wang, Oliver and Zhang, Richard and
               Owens, Andrew and Efros, Alexei A.},
  title     = {{CNN}-Generated Images Are Surprisingly Easy to Spot\ldots{}
               for Now},
  booktitle = {Proceedings of the {IEEE/CVF} Conference on Computer Vision
               and Pattern Recognition},
  pages     = {8695--8704},
  year      = {2020},
  url       = {https://arxiv.org/abs/1912.11035}
}

@inproceedings{gragnaniello2021,
  author    = {Gragnaniello, Diego and Cozzolino, Davide and Marra, Francesco
               and Poggi, Giovanni and Verdoliva, Luisa},
  title     = {Are {GAN} Generated Images Easy to Detect? {A} Critical
               Analysis of the State-of-the-Art},
  booktitle = {{IEEE} International Conference on Multimedia and Expo},
  year      = {2021},
  url       = {https://arxiv.org/abs/2104.02617}
}

@inproceedings{yu2021artificial,
  author    = {Yu, Ning and Skripniuk, Vladislav and Abdelnabi, Sahar and
               Fritz, Mario},
  title     = {Artificial Fingerprinting for Generative Models: Rooting
               Deepfake Attribution in Training Data},
  booktitle = {Proceedings of the {IEEE/CVF} International Conference on
               Computer Vision},
  pages     = {14448--14457},
  year      = {2021},
  url       = {https://arxiv.org/abs/2007.08457}
}

@techreport{c2pa2024spec,
  author      = {{Coalition for Content Provenance and Authenticity}},
  title       = {{C2PA} Technical Specification, Version 2.1},
  institution = {C2PA},
  year        = {2024},
  url         = {https://c2pa.org/specifications/}
}

@misc{openai2026c2pa,
  author  = {{OpenAI}},
  title   = {{GPT-image-2}: Our Most Capable Image Generation Model},
  year    = {2026},
  url     = {https://openai.com/blog/gpt-image-2},
  note    = {Accessed April 2026}
}

@article{shankar2024artsub,
  author    = {Sha, Pei-Ying and Lee, Kai-Cheng and Murthy, Dhiraj},
  title     = {Examining the Prevalence and Dynamics of {AI}-Generated
               Media in Art Subreddits},
  journal   = {arXiv preprint arXiv:2410.07302},
  year      = {2024},
  url       = {https://arxiv.org/abs/2410.07302}
}

@article{luo2025pixels,
  author    = {Luo, Yifan and Pierri, Francesco and Sharma, Karishma and
               Flamino, James and Szymanski, Bogdan K. and Ferrara, Emilio},
  title     = {Synthetic Politics: Prevalence, Spreaders, and Emotional
               Reception of {AI}-Generated Political Images on {X}},
  journal   = {arXiv preprint arXiv:2502.11248},
  year      = {2025},
  url       = {https://arxiv.org/abs/2502.11248}
}

@inproceedings{barrón2024ammeba,
  author    = {Horton, Christopher and others},
  title     = {{AMMeBa}: A Large-Scale Survey and Dataset of Media-Based
               Misinformation In-the-Wild},
  booktitle = {arXiv preprint arXiv:2405.11697},
  year      = {2024},
  url       = {https://arxiv.org/abs/2405.11697}
}

@misc{twitterapi2024,
  author  = {{X Corp.}},
  title   = {Twitter {API} v2 Documentation: Recent Search Endpoint},
  year    = {2024},
  url     = {https://developer.twitter.com/en/docs/twitter-api/tweets/search/api-reference/get-tweets-search-recent},
  note    = {Accessed April 2026}
}

@misc{twitter2024ailabel,
  author  = {{X Corp.}},
  title   = {Labels for {AI}-Generated Media on {X}},
  year    = {2024},
  url     = {https://help.twitter.com/en/using-x/ai-labels},
  note    = {Accessed April 2026}
}

@inproceedings{radford2021clip,
  author    = {Radford, Alec and Kim, Jong Wook and Hallacy, Chris and
               Ramesh, Aditya and Goh, Gabriel and Agarwal, Sandhini and
               Sastry, Girish and Askell, Amanda and Mishkin, Pamela and
               Clark, Jack and Krueger, Gretchen and Sutskever, Ilya},
  title     = {Learning Transferable Visual Models From Natural Language Supervision},
  booktitle = {Proceedings of the 38th International Conference on Machine Learning},
  series    = {ICML},
  year      = {2021},
  pages     = {8748--8763}
}

@article{mcinnes2018umap,
  author  = {McInnes, Leland and Healy, John and Melville, James},
  title   = {{UMAP}: Uniform Manifold Approximation and Projection for Dimension Reduction},
  journal = {arXiv preprint arXiv:1802.03426},
  year    = {2018}
}

@misc{scamai_research,
  author       = {{Scam.AI}},
  title        = {Scam.{AI} Research: Deepfake Detection Research and Datasets},
  howpublished = {\url{https://scam.ai/research}},
  year         = {2026},
  note         = {Canonical repository for Scam.AI datasets and code}
}

@misc{trustable_ai,
  author       = {{Trustable}},
  title        = {Trustable: Document and {PDF} Forgery Detection},
  howpublished = {\url{https://trustable.ai}},
  year         = {2026},
  note         = {Document and PDF forgery datasets, models, and detection tools}
}

\onecolumn
\appendix
\section{Additional Figures}

\begin{figure}[!h]
\centering
\includegraphics[width=\textwidth]{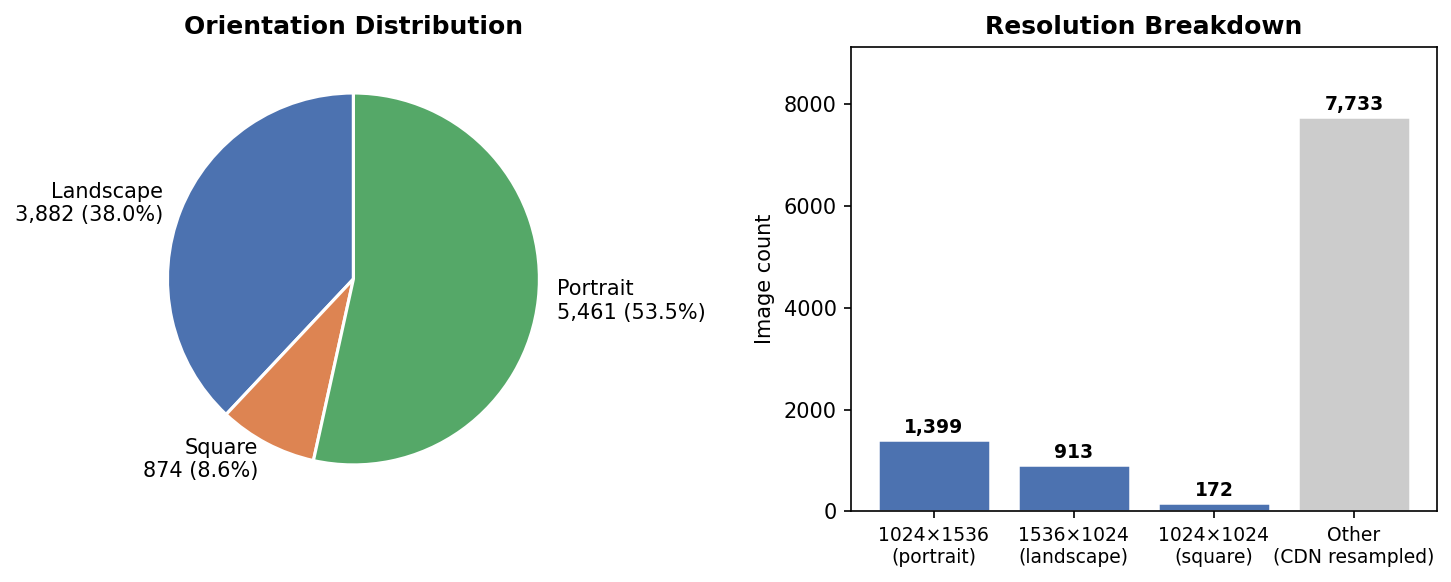}
\caption{Aspect ratio distribution (left) and native-resolution breakdown
(right) for all 10,217 confirmed images. Portrait is the plurality at 53.5\%.
``Other'' resolutions reflect Twitter CDN resampling on upload.}
\label{fig:aspect}
\end{figure}

\begin{figure}[!h]
\centering
\includegraphics[width=\textwidth]{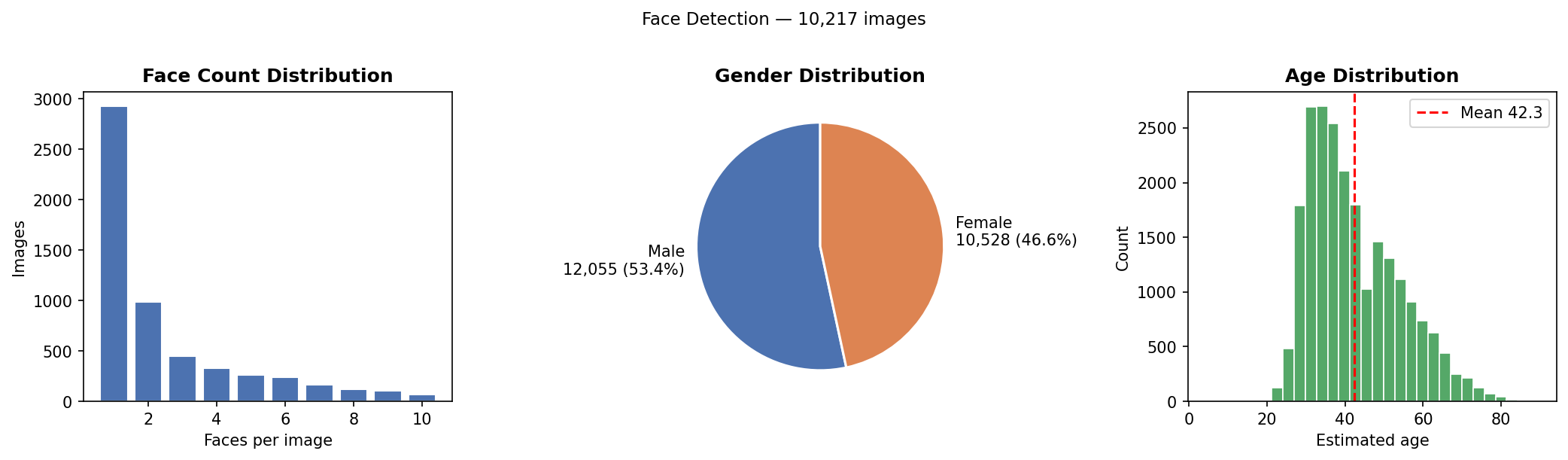}
\caption{Face count distribution (left), gender split (centre), and
estimated age histogram (right) across 10,217 confirmed images.
59.2\% of images contain at least one detected face (22,583 total faces).}
\label{fig:faces}
\end{figure}

\end{document}